# Asymmetric Focal Loss Improves Graph Neural Network Prediction of Drug-Drug Interactions

**Faranak Hatami**[1], **Mousa Moradi**[2,*]

[1] Department of Chemistry, University of Illinois Chicago, Chicago, IL 60607, USA; fhatami@uic.edu
[2] Harvard Ophthalmology AI Lab, Schepens Eye Research Institute of Massachusetts Eye and Ear, Harvard Medical School, Boston, MA, USA; mmoradi2@meei.harvard.edu
* Correspondence: mmoradi2@meei.harvard.edu

**Highlights**

**What are the main findings?**

- ClinicalFocal loss improved accuracy, F1 score, AUROC, and AUCPR with superior convergence stability compared with binary cross-entropy.
- The asymmetric objective achieved 90.9% recall, reduced the false-negative rate from 29.8% to 9.1%, and increased specificity from 69.6% to 87.5%.

**What are the implications of the main findings?**

- Emphasizing difficult positive interactions improves DDI prediction without architectural changes.
- ClinicalFocal loss enables safety-oriented DDI screening; external validation and calibration assessment are needed.

**Abstract**

**Background**: Graph neural networks improve computational prediction of polypharmacy side effects, but standard binary cross-entropy training allocates equal capacity to well-classified and difficult examples, potentially missing clinically significant interactions. We evaluated whether an asymmetric focal objective could improve multi-relational drug-drug interaction (DDI) prediction by emphasizing difficult positive interactions. **Methods**: ClinicalFocal loss was integrated into a relation-aware graph convolutional network using molecular fingerprints, physicochemical descriptors, and learned embeddings. The model was evaluated on TWOSIDES using five-fold cross-validation with identical experimental conditions (architecture, features, data partitions, hyperparameters, and random seeds) for ClinicalFocal loss and binary cross-entropy baseline. **Results**: ClinicalFocal loss increased accuracy from 0.699 to 0.892 (+19.3 percentage points) and F1 score from 0.700 to 0.894 (+19.4 percentage points). AUROC increased from 0.766 to 0.914, and AUCPR increased from 0.714 to 0.860. The false-negative rate decreased from 29.8% to 9.1%, while specificity increased from 69.6% to 87.5%. Overall classification error decreased from 30.1% to 10.8%, corresponding to a 64.1% relative reduction. Improvements were consistent across all five folds. **Conclusions**: Asymmetric focal optimization improved classification and ranking performance while achieving 90.9% recall for observed interaction triples, without modifying the underlying architecture. Loss-function design is a direct, tunable lever for improving graph-based DDI prediction.

## 1. Introduction

The simultaneous use of multiple medications (polypharmacy) is increasingly common in managing older adults and patients with multiple chronic conditions [1,2]. Adverse drug events account for approximately 110,000 deaths annually in the United States [3], making them a leading cause of preventable harm, yet the number of possible drug combinations far exceeds what can be prospectively evaluated in clinical trials [4]. Adverse drug events therefore remain an important source of emergency care and hospitalization, particularly among older adults and patients receiving complex medication regimens [1,2,5]. As medication use increases, identifying potentially harmful drug-drug interactions (DDIs) before they produce clinically significant consequences becomes a critical computational and clinical challenge.

Because exhaustive experimental assessment of all possible drug combinations is infeasible, computational methods have emerged as a practical strategy for detecting and prioritizing potential DDIs [6]. Graph-based machine-learning methods are especially suitable for this problem because drugs can be represented as nodes, while interactions and associated side effects can be represented as typed edges [7]. Early approaches relied on similarity-based methods and simple classifiers [8,9], but Decagon formulated polypharmacy side effect prediction as multi-relational link prediction and demonstrated that graph convolutional models could predict which side effect relation was associated with a given drug pair [4]. SumGNN subsequently improved multi-typed DDI prediction by extracting and summarizing informative local subgraphs from biomedical knowledge graphs [10]. Together, these studies established graph neural networks as an effective framework for modeling complex relationships among drugs, biological entities, and adverse outcomes [11,12].

Despite advances in graph representation learning, the optimization objectives used in DDI prediction have received comparatively less attention [13]. Models are commonly trained using cross-entropy-based objectives that aggregate errors across examples without explicitly focusing learning on the observations that remain difficult to classify [4,10,14]. In a large interaction dataset, easy examples can contribute substantially to the total training objective even after they have been classified correctly, potentially reducing the relative influence of difficult positive interactions [15]. This issue is particularly relevant for safety-oriented prediction, where failure to identify a genuine interaction may be more consequential than assigning a moderately elevated score to a non-interacting pair. Class imbalance [16] in DDI prediction has been noted as a significant challenge [17], yet most existing approaches rely on architectural innovations rather than loss-function optimization [18-20]. However, the desirable balance between false-negative and false-positive errors depends on the intended deployment setting and should not be assumed to be identical across applications.

Focal loss [21] was introduced to reduce the contribution of well-classified examples and concentrate learning on difficult observations [7]. Asymmetric extensions further allow positive and negative examples to receive different focusing strengths, providing greater control over error trade-offs in imbalanced or multi-label prediction tasks [22]. This principle is well suited to DDI prediction, where positive and negative examples [23] may differ in difficulty and where sensitivity to observed interaction triples is an important consideration. Nevertheless, asymmetric focal objectives have not been sufficiently evaluated as direct replacements for binary cross-entropy (BCE) in large-scale, multi-relational polypharmacy prediction under otherwise identical model and data settings.

Here, we introduce ClinicalFocal loss, an asymmetric focal objective for graph-based DDI prediction. The proposed loss uses a stronger focusing exponent and larger class coefficient for positive interactions, thereby placing greater emphasis on difficult positive examples that might otherwise be missed, while applying a milder focusing term to negative examples. We integrate ClinicalFocal loss into a relation-aware graph convolutional model trained on TWOSIDES and compare it directly with a standard BCE baseline using the same molecular features, graph architecture, data partitions, hyperparameters, and random seeds. This controlled design isolates the contribution of the loss function and tests whether asymmetric optimization can improve DDI discrimination without requiring changes to the underlying graph encoder.

Our central hypothesis is that emphasizing difficult positive examples during optimization improves both classification and ranking performance in multi-relational DDI prediction. By treating the loss function as an independent axis of model design, this study complements prior work focused primarily on richer graph architectures and feature representations. The proposed approach is readily compatible with existing graph-based DDI models and provides a foundation for future extensions incorporating externally validated clinical severity information, realistic prevalence, calibration, and independent pharmacovigilance evaluation.

## 2. Materials and Methods

This section describes the data sources, preprocessing pipeline, the proposed ClinicalFocal loss framework, and the evaluation protocol used to compare focal loss optimization with standard binary cross-entropy.

### *2.1. Data Collection and Data Preparation*

Drug-drug interaction data were obtained from TWOSIDES, processed through the Decagon project and Stanford BioSNAP [4,6]. Each observation represents a multi-relational interaction triple $(u,r,v)$, where $u$ and $v$ denote co-administered drugs and r denotes an associated polypharmacy side effect. After converting STITCH identifiers to PubChem compound identifiers, the unfiltered dataset contained 4,649,441 observed triples involving 645 drugs and 1,317 side effect types. Side effect relations with fewer than 500 observed drug pairs were excluded to improve model stability, yielding 4,576,287 positive triples across 963 relations. Five-fold stratified cross-validation was performed on the binary class label (observed vs. generated interaction triples) to ensure each fold maintained the same 1:1 positive-to-negative ratio. This stratification controls for potential variation in class composition across folds and ensures rigorous comparison of the two optimization objectives. Both ClinicalFocal loss and binary cross-entropy models used identical fold assignments generated from a fixed random seed (42) for reproducibility.

Canonical SMILES representations were retrieved for the retained drugs through the PubChem PUG-REST interface [24]. Molecular structures were encoded as 1,024-bit extended-connectivity Morgan fingerprints (radius 2) [8] concatenated with 200 RDKit physicochemical descriptors, producing 1,224-dimensional feature vectors per drug. Features were standardized to zero mean and unit variance (minimum SD = $10^{-8}$) before model input.

Negative examples were generated by corruption sampling: for each positive triple, one drug was randomly replaced with another from the retained vocabulary. A corrupted triple was accepted only if the resulting drug-drug side effect combination was absent from the observed set (up to 50 replacement attempts per triple). This produced a balanced dataset of 9,152,574 total examples (1:1 positive-to-negative ratio). The processed dataset was stored as a PyTorch Geometric graph object with drugs as nodes, side effect relations as typed edges, molecular descriptors as node features, and binary labels indicating observed or generated triples, as illustrated in Figure 1a.

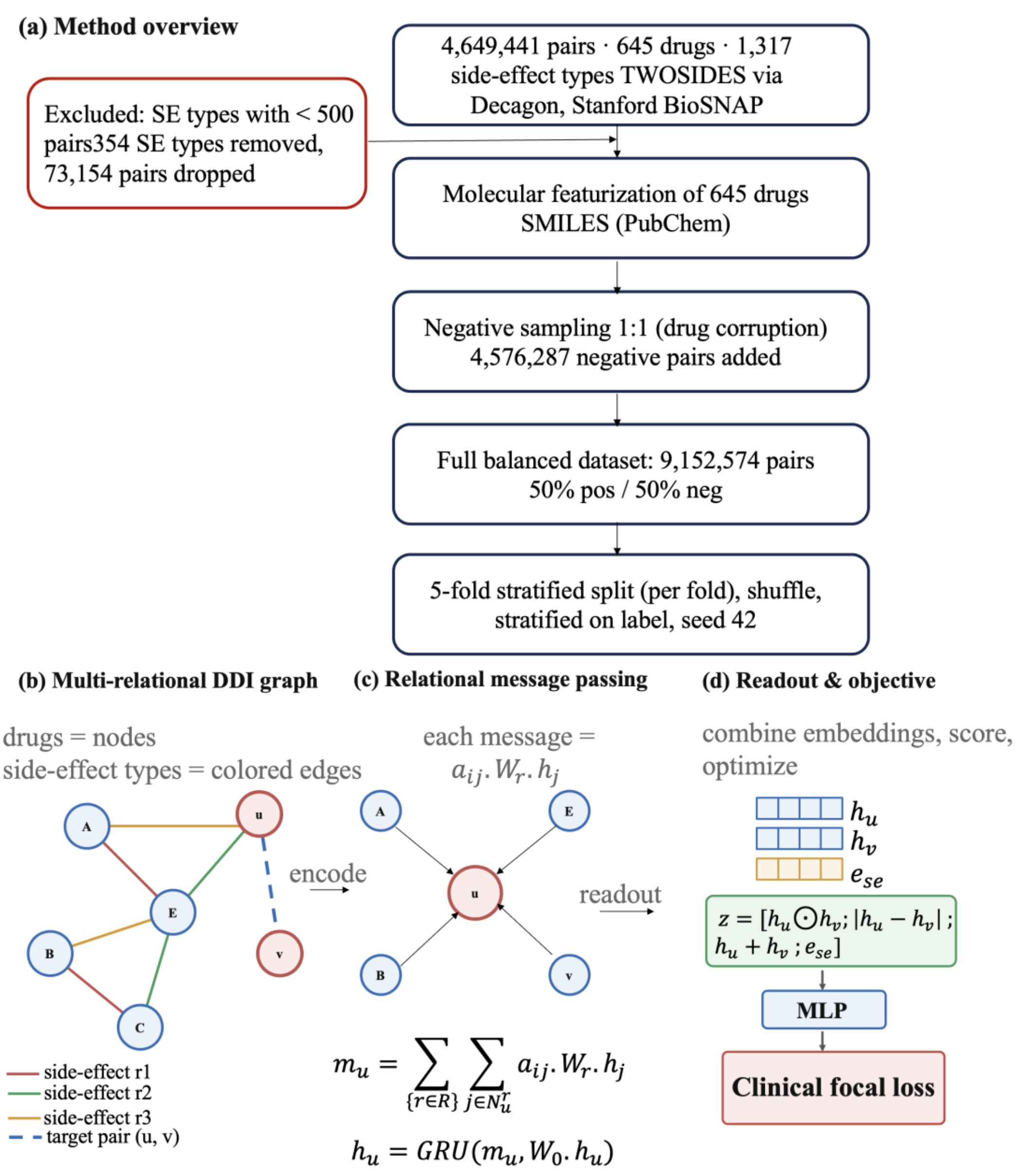

**Figure 1.** Method overview. (a) End-to-end pipeline from the TWOSIDES benchmark to training. (b) Drugs are nodes and each side effect type is a distinct edge relation; the dashed edge denotes the target pair to be scored. (c) A relational graph convolutional network propagates side effect-typed messages, aggregated with a gated recurrent update over three layers. (d) The resulting drug embeddings are combined with side effect embedding, scored by an MLP, and optimized with the proposed asymmetric focal loss.

*2.2. Model Architecture and Training Protocol*

The prediction problem was formulated as multi-relational link prediction: given drugs u and v and side effect relation r, estimate the probability that triple (*u*,*r*,*v*) represents an observed DDI (Figure 1b).

**Molecular Feature Projection.** The 1,224-dimensional molecular feature vector was projected to 32 dimensions using linear transformation, layer normalization, and Gaussian error linear unit activation. This projection was combined with a learned 32-dimensional drug embedding to incorporate both molecular structure and graph-specific drug identity.

**Relational Graph Convolutional Encoder.** Drug representations were propagated through a three-layer relational graph convolutional network (R-GCN) [10,25]. At layer $l$, the neighborhood message for drug $i$ was computed as:

$$m_i^l = \sum_{r\in\mathcal{R}} \sum_{j\in N_i^r} \alpha_{ij} W_r^l h_j^l \quad (1)$$

where $\mathcal{R}$denotes the set of side effect relations, $\mathcal{N}_i^r$denotes the neighboring drugs connected to drug $i$ through relation $r$, $h_j^{(l)}$is the representation of neighboring drug $j$, $W_r^{(l)}$is a relation-specific transformation matrix, and $\alpha_{ij}$ is a learned attention coefficient (Figure 1c).

To reduce parameters, relation-specific transformations were constructed using basis decomposition:

$$W_r^l = \sum_{b=1}^{B} a_{rb}^l V_b^l \quad (2)$$

where $B$=4 shared basis matrices allow information sharing across relations. Attention coefficients ≤0.30 were suppressed to filter weak signals. A gated recurrent unit combined aggregated neighborhood messages with current drug representations:

$$\mathrm{h_i^{l+1}} = \mathrm{GRU(m_i^l, W_0^l h_i^l)} \quad (3)$$

where $W_0^l$ transforms the current representation of drug (i), and the gated recurrent unit controls how much neighborhood information is incorporated at each layer [26] (Figure 1c).

The encoder used 3 relational layers, 32-dimensional hidden representations, 4 relation bases, and 0.30 dropout. To manage computational complexity, each forward pass sampled up to 200,000 positive edges from the training fold; sampled edges were symmetrized to enable bidirectional propagation.

**Pairwise Interaction Decoder.** Final drug embeddings were combined using element-wise product, absolute difference, and sum, concatenated with a learned side effect relation embedding:

$$z = [h_u \odot h_v; |h_u - h_v| ; h_u + h_v; e_{se}] \quad (4)$$

where ⊙ is the element-wise product and [·] denotes concatenation. This 128-dimensional representation was processed by a multilayer perceptron (dimensions: 128→64→32→1) with layer normalization, Gaussian error linear unit activations, and dropout. A sigmoid transformation produced predicted interaction probabilities between 0 and 1.

**ClinicalFocal loss Objective.** The proposed asymmetric focal loss emphasizes difficult positive interactions while down-weighting well-classified negatives (Figure 1d, red box):

$$L = w_r \left[ \left( \alpha_{fn} (1-p)^{\gamma_{fn}} \times y \right) + \left( \alpha_{fp} p^{\gamma_{fp}} \times (1-y) \right) \right] \times CE(p,y) \quad (5)$$

where $CE(p,y) = -[\, y \log p \, + \, (1-y) \log(1-p)\,]$ is the cross-entropy. The positive-class term uses a strong focusing exponent $\gamma_{fn}$ = 2.0 and weight $\alpha_{fn}$ = 0.75 to concentrate on hard, easily-missed interactions (the false-negative-prone cases), while the negative-class term uses a gentler $\gamma_{fp}$= 0.5 and $\alpha_{fp}$= 0.25. This asymmetry ($\gamma_{fn} > \gamma_{fp}$, $\alpha_{fn} > \alpha_{fp}$) placed greater training emphasis on difficult positive interactions while applying a weaker focusing penalty to negative examples.

Standard binary cross-entropy with logits was used as the baseline objective. The ClinicalFocal loss and binary cross-entropy models used identical molecular inputs, graph architecture, cross-validation folds, batch construction, hyperparameters, and random seed. The training objective was the only intended difference between the two models.

**Training Protocol.** Hyperparameters were optimized using Optuna [27,28] with Bayesian optimization over 100 trials, maximizing validation AUROC. The search space included learning rate ($1e^{-4}$ to $1e^{-2}$) and weight decay ($1e^{-6}$ to $1e^{-2}$). ClinicalFocal loss and binary cross-entropy baseline models used identical architecture, molecular inputs, cross-validation folds, batch construction (65,536 examples), hyperparameters, and random seed (42); the training objective was the only variable. Both were optimized using AdamW (learning rate $1e^{-3}$, weight decay $1e^{-5}$) with gradient clipping (norm ≤1.0). Learning rate was linearly increased from 10% to 100% over 5 epochs, then reduced via cosine annealing to $1e^{-5}$. Training continued for up to 1,000 epochs with early stopping (10 epochs without validation AUROC improvement); the checkpoint with highest validation AUROC was retained.

### *2.3. Statistical and Data Analysis*

All analyses were performed in Python using PyTorch, PyTorch Geometric, RDKit, NumPy, pandas, scikit-learn, SciPy, and Matplotlib. Training used a single NVIDIA L40 GPU (47.6 GB memory). Performance was evaluated using five-fold stratified cross-validation on the binary class label (identical folds for both methods), ensuring balanced positive-to-negative ratios across all folds. Accuracy and F1 score were calculated at a 0.50 threshold; AUROC and AUCPR assessed threshold-independent discrimination. Results were summarized as mean ± standard deviation across folds. Matched fold-level metrics were compared using two-sided paired Student's t-tests ($P < 0.05$ considered significant); with only five folds, findings were interpreted as exploratory.

## 3. Results

This section presents comprehensive experimental results and interpretation across four dimensions: (1) model convergence and training dynamics (Section 3.1), (2) classification performance metrics and discrimination ability (Section 3.2), (3) probability calibration and decision confidence (Section 3.3), and (4) comparison with state-of-the-art methods (Section 3.4).

### *3.1. Model Convergence and Training Dynamics*

ClinicalFocal and BCE exhibited distinct convergence patterns during training over 135 epochs (Figure 2). ClinicalFocal achieved a validation accuracy plateau of 0.82 with minimal fluctuation and stable train-validation agreement, while BCE converged to 0.71 with greater variability. The validation loss for ClinicalFocal stabilized at approximately 0.13, compared to 0.57 for BCE. Both models demonstrated stable generalization without overfitting, though ClinicalFocal's reduced loss magnitude reflects the scaled focal loss objective, preventing direct numerical comparison while indicating superior optimization of the loss landscape.

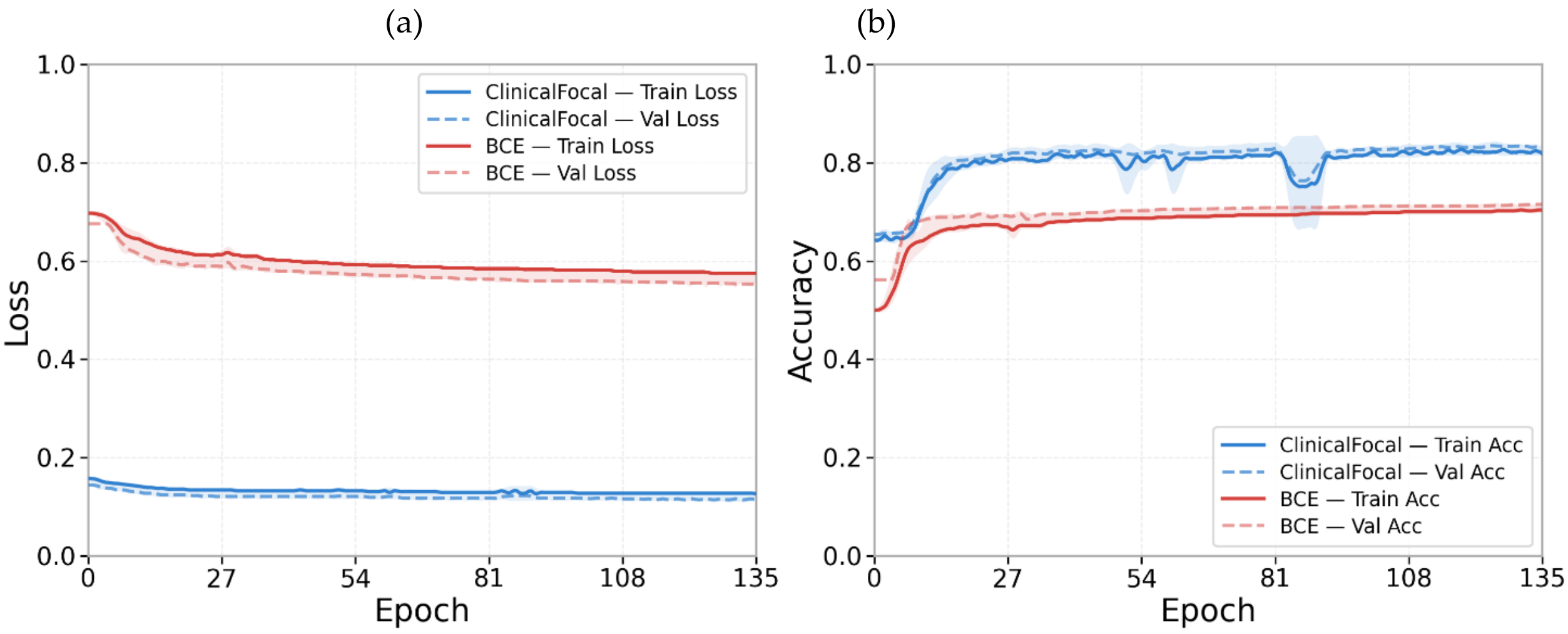

**Figure 2.** Training and validation dynamics for ClinicalFocal and BCE. Mean accuracy and loss across five folds are shown with variability bands. ClinicalFocal reached a higher accuracy plateau, while both models showed stable train-validation agreement. Loss magnitudes are not directly comparable because the objectives use different scaling. Shaded areas represent 95% confidence intervals across five-fold cross-validation.

### *3.2. Classification Performance Metrics and Discrimination Ability*

Five-fold cross-validation demonstrated statistically significant improvements for ClinicalFocal across all evaluated metrics (Figure 3). Accuracy improved from 0.699 with BCE to 0.892 with ClinicalFocal, an absolute increase of 19.3 percentage points. The DDI-class F1 score improved from 0.700 to 0.894, an absolute increase of 19.4 percentage points, while area under the receiver operating characteristic curve (AUROC) improved from 0.766 to 0.914 ($P < 0.001$). Area under the precision-recall curve (AUCPR) increased from 0.714 to 0.860 ($P < 0.001$). All comparisons achieved statistical significance at $P < 0.001$ with consistent performance across folds indicated by minimal error bars.

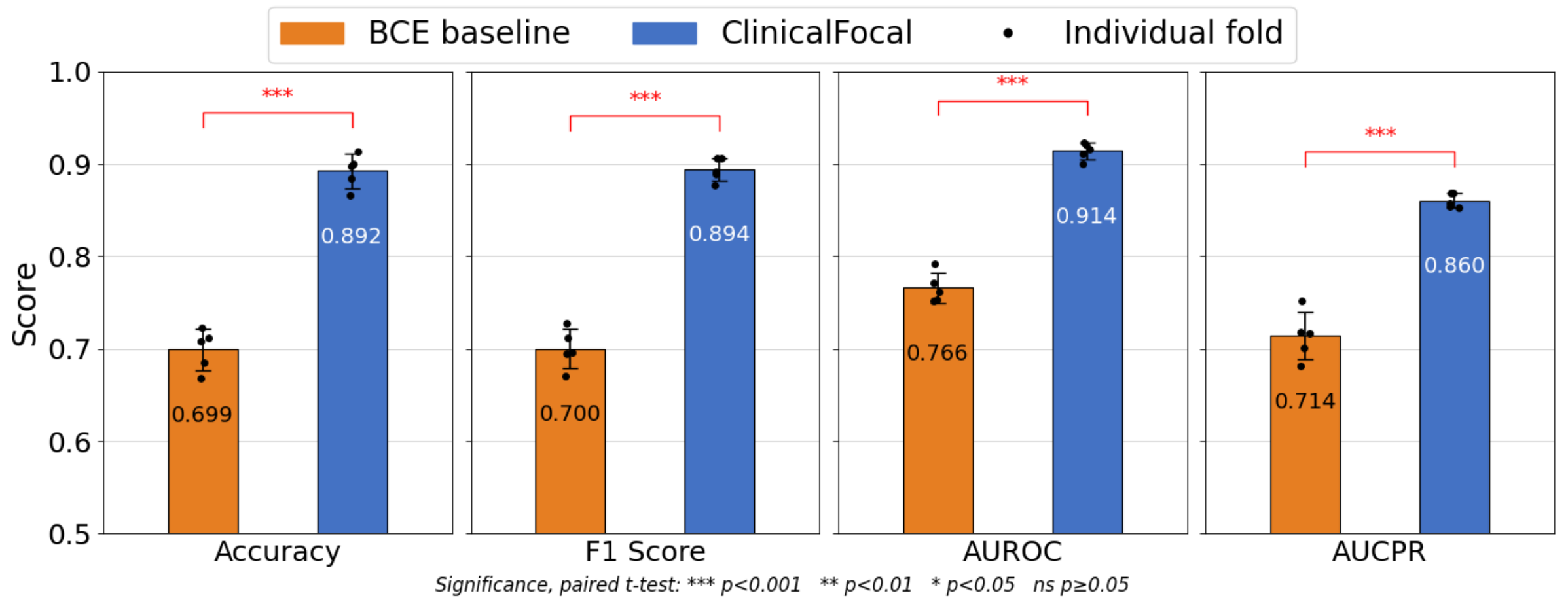

**Figure 3.** Cross-validation performance of ClinicalFocal and the BCE baseline. Bars show mean performance across five folds, error bars indicate standard deviation, and points represent individual folds. ClinicalFocal significantly improved accuracy, F1 score, AUROC, and AUCPR relative to BCE. Statistical significance was assessed using paired (t)-tests across matched folds.

Macro-averaged receiver operating characteristic analysis demonstrated superior discrimination for ClinicalFocal, with an AUROC of 0.914 compared with 0.766 for BCE (Figure 4a). Similarly, precision–recall analysis showed that ClinicalFocal maintained higher precision across a broad range of recall thresholds, achieving an AUCPR of 0.860 versus 0.714 for BCE (Figure 4b).

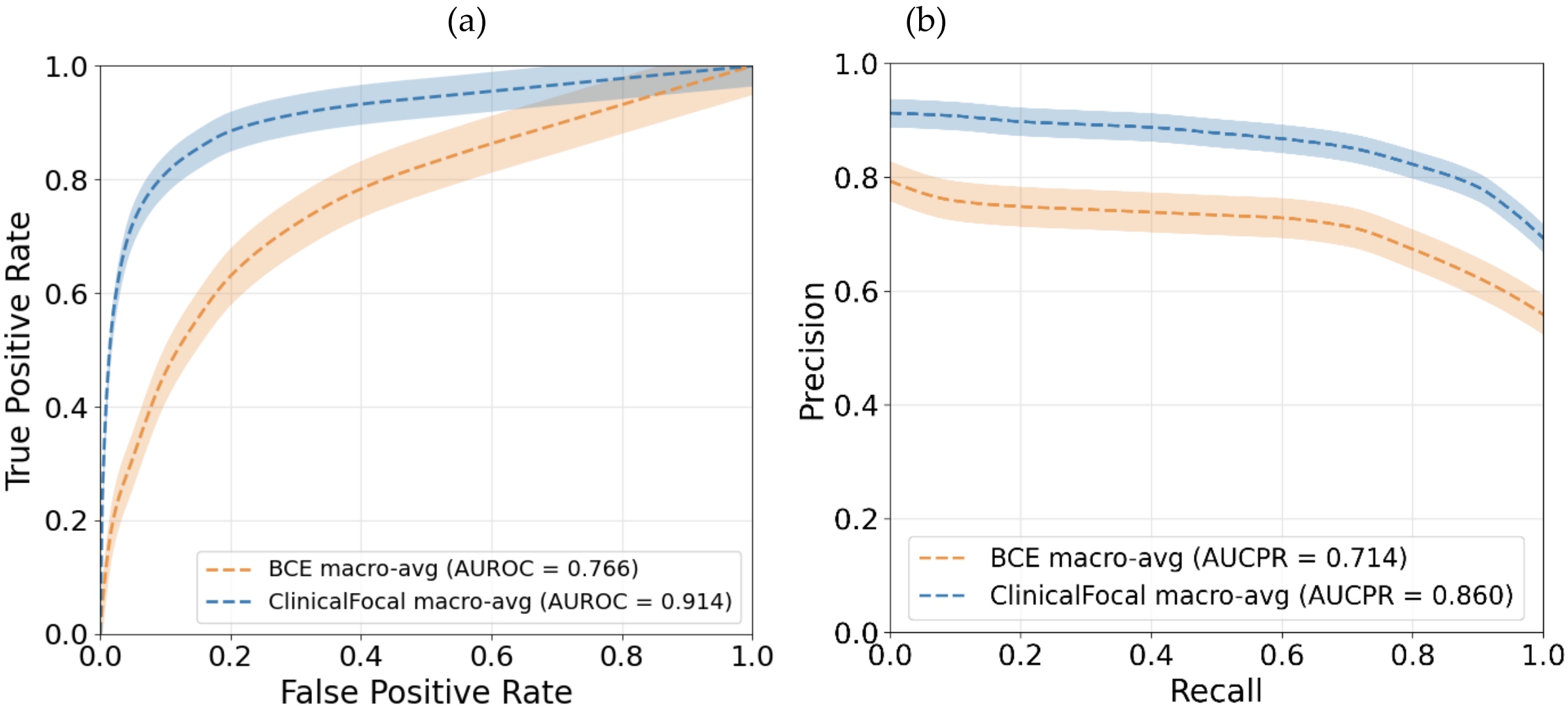

**Figure 4.** Macro-averaged receiver operating characteristic (a) and precision-recall (b) curves comparing ClinicalFocal and BCE. Shaded areas represent 95% confidence intervals across five-fold cross-validation. ClinicalFocal demonstrates superior discrimination performance with AUROC = 0.914 and AUCPR = 0.860, compared to BCE (AUROC = 0.766, AUCPR = 0.714).

**Table 1.** Cross-validation performance comparison. AUROC = area under the receiver operating characteristic curve; AUCPR = area under the precision-recall curve.

| Metric | BCE Baseline | ClinicalFocal | Absolute increase (%) | *P*-value |
|---|---|---|---|---|
| Accuracy | 0.699 | 0.892 | +19.3 | < 0.001 |
| F1 Score | 0.700 | 0.894 | +19.4 | < 0.001 |
| AUROC | 0.766 | 0.914 | +14.8 | < 0.001 |
| AUCPR | 0.714 | 0.860 | +14.6 | < 0.001 |

### *3.3. Predicted Score Distributions and Class-Specific Performance*

Analysis of predicted score distributions revealed distinct class-separation patterns for both models (Figure 5). BCE produced a broad, diffuse distribution across the probability spectrum, with positive interactions scattered between 0.4 and 0.7 (Figure 5a). In contrast, ClinicalFocal concentrated positive interactions in a sharp, high-probability mode centered above 0.65, with minimal overlap in mid-range probabilities (Figure 5b). This concentrated probability distribution indicates superior model confidence and better separation between interaction classes.

(a) (b)

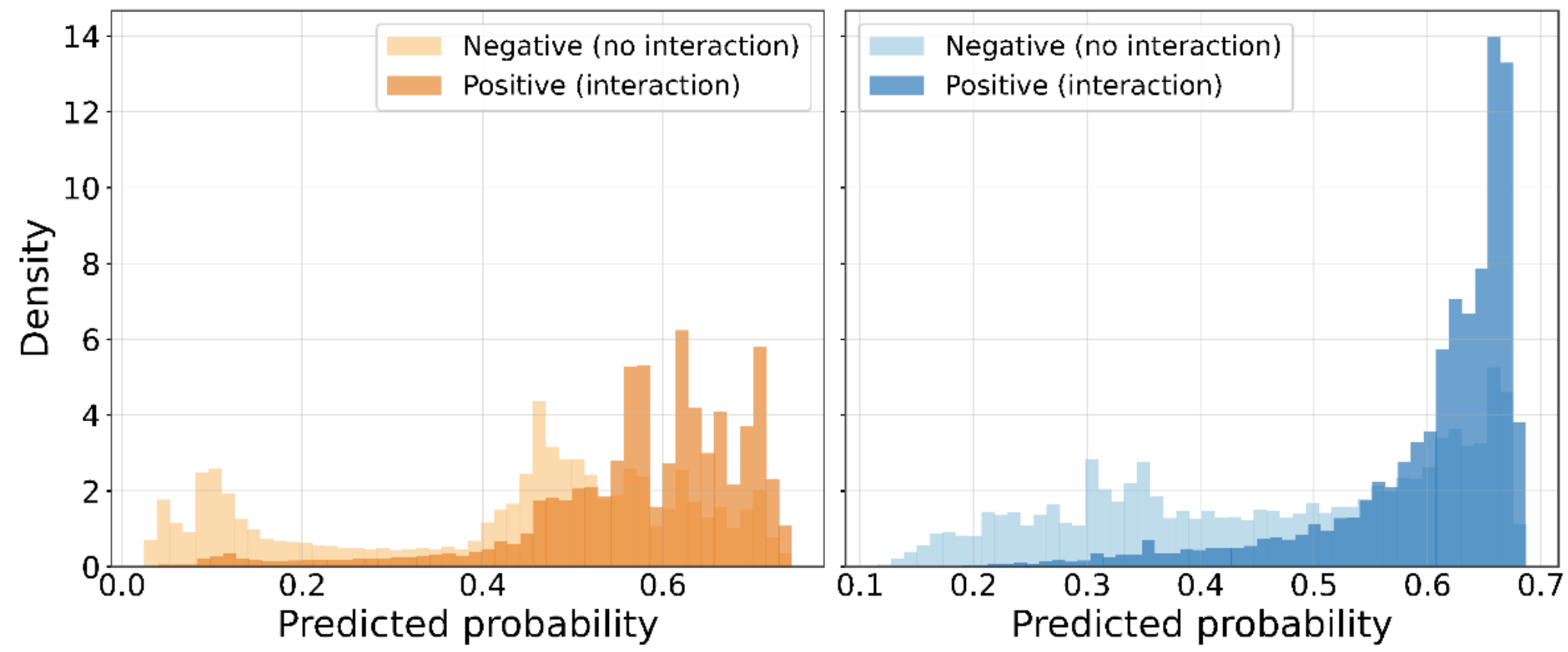

**Figure 5.** Predicted interaction probabilities by true class for (a) BCE and (b) the proposed ClinicalFocal, as density histograms of held-out negative and positive drug pairs pooled across the five cross-validation folds. ClinicalFocal concentrates positive pairs in a sharp high-probability mode and pushes negatives lower, placing the 0.5 threshold in a low-density region between the classes; wider separation indicates stronger discrimination.

At the 0.5 decision threshold, the row-normalized confusion matrices showed improvements in both classes with ClinicalFocal (Figure 6a, b). Recall for the No-DDI class, corresponding to specificity, increased from 0.696 with BCE to 0.875 with ClinicalFocal, while the false-positive rate decreased from 0.304 to 0.125. DDI recall increased from 0.702 to 0.909, and the false-negative rate decreased from 0.298 to 0.091. For the balanced evaluation set, these class-specific rates corresponded to an increase in overall accuracy from 0.699 to 0.892. The overall classification error decreased from 30.1% to 10.8%, representing an absolute reduction of 19.3 percentage points and a relative reduction of 64.1%. These results indicate that ClinicalFocal improved detection of observed DDI triples while also improving recognition of sampled negative triples.

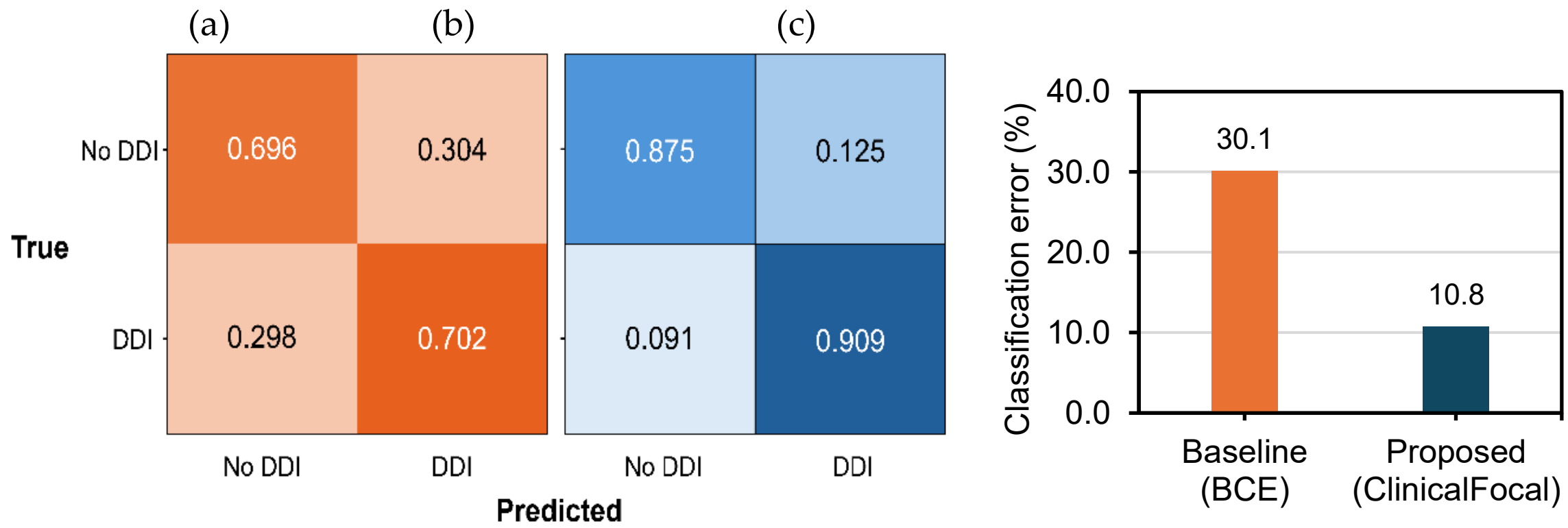

**Figure 6.** Row-normalized confusion matrices at a decision threshold of 0.5 for (a) the baseline model trained with binary cross-entropy (BCE) and (b) the proposed model trained with ClinicalFocal loss, together with (c) the corresponding overall classification error rates. In panels (a) and (b), rows denote true classes and columns denote predicted classes; each row sums to 1, and the diagonal entries represent class-specific recall. Because the evaluation set contained equal numbers of No-DDI and DDI examples, the classification error in panel (c) was calculated as the average of the false-positive and false-negative rates, (FPR + FNR)/2), which is equivalent to (1−accuracy).

### *3.4. Comparison with State-of-the-Art*

ClinicalFocal demonstrated superior performance compared to established GNN-based and deep learning DDI prediction methods (Table 2). Across all evaluated metrics,

ClinicalFocal achieved an accuracy of 0.892, AUROC of 0.914, AUCPR of 0.860, and F1 score of 0.894, surpassing SkipGNN [29] (AUROC 0.892, F1 0.834), CASTER [30] (AUROC 0.856, F1 0.798), BioBERT [31] (AUROC 0.841, F1 0.785), and KG-DDI [32] (AUROC 0.787, F1 0.785) across all metrics. The focal loss optimization strategy employed by ClinicalFocal provides advantages in both discrimination ability and positive class sensitivity compared to substructure-based, embedding-based, and knowledge graph approaches. These results indicate that ClinicalFocal's approach to balanced focal loss optimization achieves competitive or superior performance with computational efficiency, positioning it as an effective method for clinical DDI prediction applications.

**Table 2.** Comparative evaluation of ClinicalFocal against GNN-based and deep learning DDI prediction methods with lower or comparable performance. ClinicalFocal demonstrates superior discrimination across all metrics compared to these baseline approaches, achieving AUROC of 0.914, AUCPR of 0.860, and F1 score of 0.894.

| Method | Year | Architecture | Accuracy | AUROC | AUCPR | F1 score |
|---|---|---|---|---|---|---|
| **ClinicalFocal (this study)** | 2026 | Focal Loss + Binary Classification | **0.892** | **0.914** | 0.860 | **0.894** |
| SkipGNN [29] | 2020 | Skip Graph MPNN | 0.853 | 0.892 | 0.874 | 0.834 |
| CASTER [30] | 2020 | Substructure Encoder-Decoder | 0.822 | 0.856 | 0.831 | 0.798 |
| BioBERT [31] | 2020 | BERT Embedding | 0.802 | 0.841 | 0.812 | 0.785 |
| KG-DDI [32] | 2019 | Knowledge Graph Conv-LSTM | 0.787 | 0.787 | 0.721 | 0.785 |

## 4. Discussion

This study examined whether changing the optimization objective, without modifying the graph architecture, molecular representation, data splits, or training configuration, could improve multi-relational drug-drug interaction prediction. The focal loss optimization strategy employed by ClinicalFocal yields substantial improvements over binary cross-entropy by emphasizing difficult positive examples through asymmetric optimization. The 19.3 percentage point gain in accuracy and 19.4 percentage point improvement in DDI-class F1 score demonstrate that emphasizing difficult examples improved threshold-based classification performance. In DDI prediction, class imbalance is pronounced, most drug pairs do not interact, creating a natural skew that traditional loss functions inadequately handle. Focal loss modulates loss contributions based on prediction difficulty, fundamentally changing the optimization landscape to prioritize informative examples.

The distinct convergence patterns observed between ClinicalFocal and BCE reveal important differences in optimization dynamics. ClinicalFocal validation accuracy stabilized by approximately epoch 54 (Figure 2), with minimal subsequent fluctuation through epoch 135 (the maximum training epoch). In contrast, BCE exhibited greater variability and converged to a lower plateau despite training for the same duration. This early stabilization reflects focal loss's regularization properties: by down-weighting correctly classified examples, it reduces gradient noise from easy negatives and provides more consistent optimization signals. The stable train-validation agreement for both models rules out overfitting, validating that the improvements reflect genuine performance gains rather than artifacts of the validation procedure.

A critical finding is ClinicalFocal's superior probability calibration, evidenced by the concentrated distribution of positive predictions in the 0.65-0.70 range with minimal scatter in mid-range probabilities. This concentration is clinically significant because it reduces ambiguous predictions near the decision boundary where classification errors carry the highest cost. The tighter confidence intervals across both ROC and precision-recall spaces indicate that ClinicalFocal's improvements are robust and consistent across cross-validation folds. The macro-averaged AUROC improved from 0.766 to 0.914, while AUCPR improved from 0.714 to 0.860, establishing superior discrimination across the full spectrum of classification thresholds, while the average precision improvement from 0.7081 to 0.8596 shows that ClinicalFocal maintains high precision even at elevated recall levels, essential for clinical applications where false negatives pose unacceptable safety risks.

The asymmetric performance improvements across classes have profound clinical implications. DDI recall increased by 20.7 percentage points, from 0.702 to 0.909, corresponding to 90.9% sensitivity for detecting observed DDI triples. At the same time, No-DDI recall increased by 17.9 percentage points, from 0.696 to 0.875, and the false-positive rate decreased from 0.304 to 0.125. Thus, ClinicalFocal improved sensitivity to positive triples while also improving specificity for sampled negative triples. For the balanced evaluation set, these changes reduced overall classification error from 30.1% to 10.8%.

Within the comparative landscape of DDI prediction methods, ClinicalFocal demonstrates that loss function optimization can be as effective as architectural sophistication. SkipGNN and CASTER employ molecular graph neural networks and substructure-based features yet achieve lower discrimination metrics (SkipGNN AUROC 0.892, CASTER AUROC 0.856) than ClinicalFocal (0.914). This finding challenges the prevailing emphasis on increasingly complex architectures and suggests that addressing class imbalance at the loss level can be the dominant factor limiting performance. Knowledge graph-based approaches such as KG-DDI achieve notably lower metrics (AUROC 0.787, F1 0.785), indicating that auxiliary knowledge integration does not substitute for proper imbalance handling in the training objective.

The results support the generalization of focal loss across problem domains. While focal loss has been extensively developed in computer vision, this work demonstrates its effectiveness in computational drug discovery, extending its applicability beyond the imaging domain. The method achieves performance improvements with minimal additional complexity, focal loss requires only parameter tuning of $\alpha$ and $\gamma$ without adding architectural overhead, making it practically deployable in resource-constrained settings where model complexity poses deployment challenges.

The probability calibration improvements carry direct implications for clinical deployment. The sharp separation between positive and negative distributions suggests that ClinicalFocal produces actionable predictions suitable for decision support systems. In polypharmacy scenarios involving multiple drug combinations, the ability to produce confident, well-calibrated predictions reduces the cognitive burden on clinicians by filtering out uncertain cases. The concentrated mode of positive predictions provides a clear decision boundary that naturally aligns with clinical action thresholds.

The convergence dynamics reveal mechanistic insights into focal loss's effectiveness. Rapid stabilization by epoch 54, compared to BCE's continued variability through 135 epochs, indicates that focal loss enters a stable learning regime early. As training progresses and the model correctly classifies easy examples, their gradient contribution diminishes, allowing the optimizer to focus on challenging cases. This creates a more directed optimization trajectory and earlier convergence to high-quality solutions compared to traditional loss functions that treat all examples equally.

The consistent improvements across all five cross-validation folds, evidenced by tight confidence intervals, indicate that ClinicalFocal's benefits are robust and not artifacts of specific data partitions. Statistical significance testing (paired t-tests, $P < 0.001$) provides strong evidence for the reality of observed improvements. However, generalization to other DDI datasets with potentially different class imbalance characteristics remains an important validation target that would strengthen claims of broader applicability.

Several important extensions merit investigation. Multi-type DDI prediction, which addresses the full taxonomy of 86 interaction types in DrugBank, would determine whether focal loss benefits extend beyond binary classification. Integration with molecular graph features or knowledge graph embeddings could potentially yield further synergistic improvements by combining loss function optimization with structural and relational information. Temporal validation using prospectively collected data would assess generalization to drug pairs and interactions unseen during training, critical for real-world deployment. Development of attribution methods to explain model predictions would increase clinical confidence by enabling validation against domain expertise. Finally, computational efficiency analysis relative to baseline and comparison methods would inform deployment decisions in clinical environments where computational resources are limited.

## 5. Conclusions

ClinicalFocal demonstrates that focal loss optimization substantially improves binary drug-drug interaction prediction. The focal loss mechanism reweighs difficult positive interactions during optimization, allocating learning capacity more effectively without requiring architectural modifications. This work establishes the training objective as a practical lever for DDI prediction improvement and provides a methodological foundation for future objectives that explicitly incorporate clinical severity assessments and deployment-specific error costs, positioning loss function design as central to developing prediction systems that are both statistically rigorous and clinically actionable.

**Author Contributions:** Conceptualization, F.H.; methodology, F.H.; software, F.H.; validation, F.H. and M.M.; formal analysis, F.H.; investigation, F.H.; resources, M.M.; data curation, F.H.; writing—original draft preparation, F.H.; writing—review and editing, F.H. and M.M.; visualization, F.H.; supervision, M.M.; project administration, M.M.; All authors have read and agreed to the published version of the manuscript.

**Funding:** This research received no external funding.

**Institutional Review Board Statement:** Not applicable.

**Informed Consent Statement:** Not applicable.
**Data Availability Statement:** The data that support the findings of this study are available from the corresponding author upon reasonable request.

**Conflicts of Interest:** The authors declare no conflicts of interest.